\documentclass[letterpaper, 10 pt, conference]{ieeeconf}  

\IEEEoverridecommandlockouts                              

\usepackage{graphics} 
\usepackage{epsfig} 
\usepackage{mathptmx} 
\usepackage{times} 
\usepackage{amsmath} 
\usepackage{amssymb}  
\usepackage{color}
\usepackage{multicol}
\usepackage{multirow}
\usepackage[table,xcdraw]{xcolor}
\usepackage{subcaption} 
\usepackage{caption}
\usepackage{dblfloatfix}

\usepackage[pagebackref=true,breaklinks=true,letterpaper=true,colorlinks,bookmarks=false]{hyperref}
\usepackage[normalem]{ulem}
\usepackage{algorithm}
\usepackage[noend]{algpseudocode}
\useunder{\uline}{\ul}{}
\newcommand*\rot{\rotatebox{90}}

\title{\LARGE \bf
Text3DAug - Prompted Instance Augmentation for LiDAR Perception
}

\author{
Laurenz Reichardt$^{1,*}$, Luca Uhr$^{1,*}$ and Oliver Wasenm\"uller$^{1}$\\
$^{1}$Mannheim University of Applied Sciences, Germany \\
{\tt\small \{l.reichardt, l.uhr, o.wasenmueller\}@hs-mannheim.de}
\thanks{* Equal Contribution}
}

\begin{document}

\maketitle
\thispagestyle{empty}
\pagestyle{empty}

\begin{abstract}

LiDAR data of urban scenarios poses unique challenges, such as heterogeneous characteristics and inherent class imbalance. Therefore, large-scale datasets are necessary to apply deep learning methods. Instance augmentation has emerged as an efficient method to increase dataset diversity. However, current methods require the time-consuming curation of 3D models or costly manual data annotation.
To overcome these limitations, we propose Text3DAug, a novel approach leveraging generative models for instance augmentation.
Text3DAug does not depend on labeled data and is the first of its kind to generate instances and annotations from text.
This allows for a fully automated pipeline, eliminating the need for manual effort in practical applications.
Additionally, Text3DAug is sensor agnostic and can be applied regardless of the LiDAR sensor used.
Comprehensive experimental analysis on LiDAR segmentation, detection and novel class discovery demonstrates that Text3DAug is effective in supplementing existing methods or as a standalone method, performing on par or better than established methods, however while overcoming their specific drawbacks.
The code is publicly available.
\footnote{\scriptsize\url{http://github.com/CeMOS-IS/Text3DAug-Augmentation}}
\end{abstract}
\section{INTRODUCTION}

LiDAR sensors enable the 3D perception of environments and are crucial for applications such as autonomous navigation, robotics, mapping and various industrial applications.
While deep learning applications have become the de facto standard for many tasks such as LiDAR detection and segmentation, the data still poses unique challenges.

Firstly, LiDAR data is heterogeneous, with characteristics highly dependend on the sensor. Point cloud structure and distribution varies with the number of scanlines, field of view, rotation frequency, mounting height, etc.
This leads to a significant decline in performance, when deep learning methods trained on data from one sensor are applied to data from another sensor.
The magnitude of this so called sensor domain gap is unique to 3D point clouds with ongoing research on how to pre-train networks on different datasets or enable multi-dataset training \cite{Ponderv2, PPT}.

Secondly, data-imbalance is inherent to LiDAR point clouds, due to multiple factors.
In the case of urban scenarios, large objects such as buildings are represented by more points compared to smaller objects or individuals. Due to the radiating alignment of the vertical LiDAR scanlines, point cloud density decreases with increased object distance, meaning that small objects are represented by few or no points beyond a certain distance.
This results in adverse effects on network performance for long range perception \cite{LRPD}.
This is exasperated by the fact that some objects, especially road participants such as motorcycles, are rare.
Such factors result in data-imbalance in large scale datasets.
For example, the SemanticKITTI dataset contains building points exceeding those representing people by a factor $709$ and for motorcyclist by a factor of $16,205$.

\begin{figure}[t!]
  \centering
  \includegraphics[width=1.0\columnwidth]{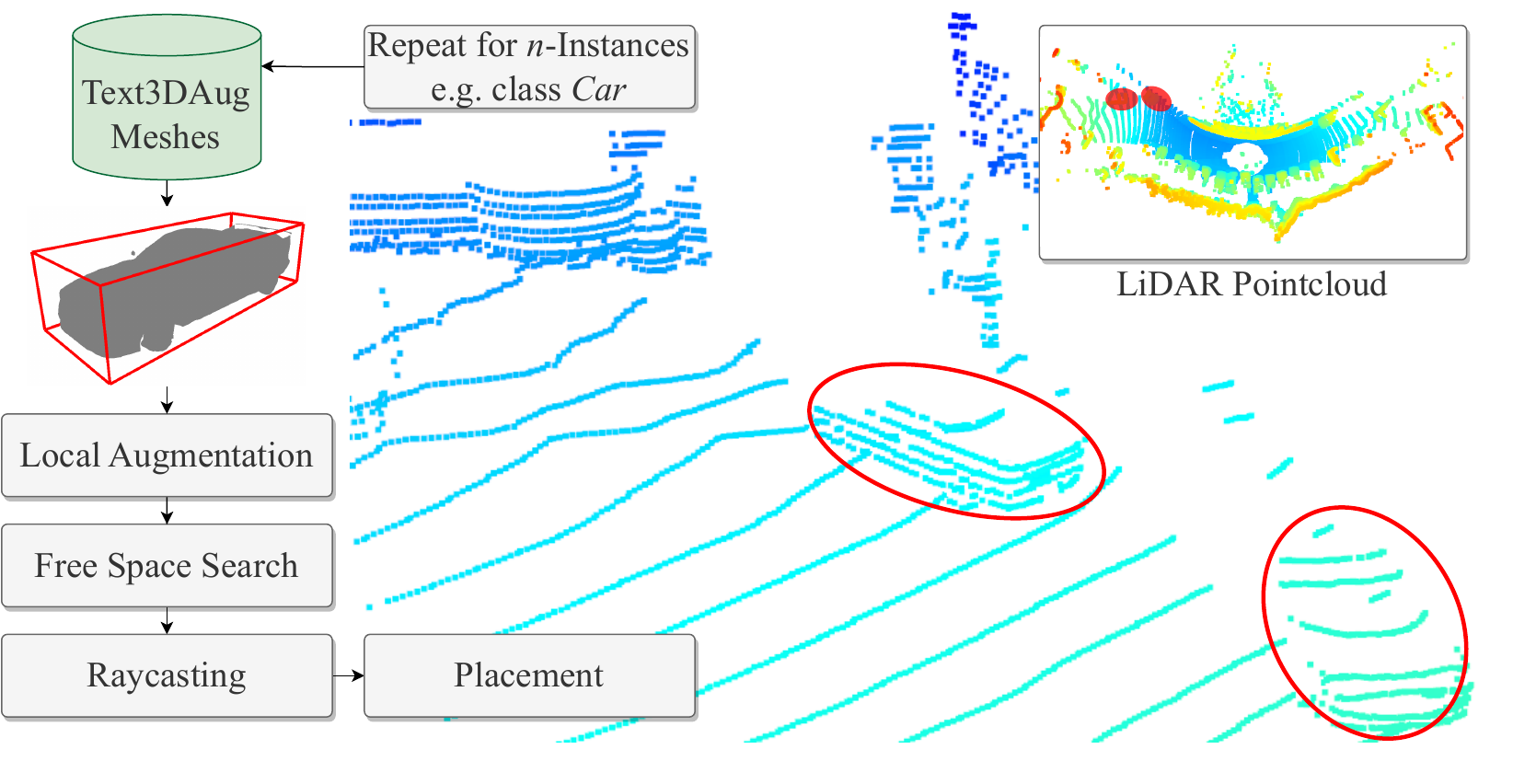}
  \caption{
The Text3DAug augmentation pipeline. We prompt our instance generation engine in order to create and annotate meshes for desired classes. These are then realistically placed and rendered in LiDAR point clouds as instances.
  }
  \label{fig:Instance_Placement}
\end{figure}

These challenges impose the need for large scale and diverse datasets in order to apply deep learning methods to LiDAR data, in order to obtain sufficient points for all classes.
Data augmentation is a standard technique to artificially increase data diversity and in the context of LiDAR scans, instance augmentation has emerged as an effective approach to tackle data-imbalance.
Specifically, training data is enriched by "cut and pasting" object instances (e.g. road participants for SemanticKITTI) from different scans.
However, the practical application of this concept has extensive requirements.
Creating instance cut-outs necessitate semantic and instance labels, however labeling point clouds is significantly more time-consuming when compared to image data due to additional dimensions involved \cite{SALT_labeling}.

The labeled data use for cut-out instances also has to exhibit sufficient objects of the desired class, which can be challenging due to data-imbalance, possibly requiring further data collection.
Moreover, these instances retain the point structure and remission values specific to their original position, LiDAR sensor and possible occlusion.
Additional factors such as the sensor domain gap, different semantic classes, or missing instance labels, mean that objects from other datasets can rarely be employed.

In this work, we tackle the above mentioned limitations, presenting Text3DAug as the first fully-automated and label-free instance augmentation method (see Figure \ref{fig:Instance_Placement}).
We aim to establish Text3DAug as a practical alternative or addition to existing methods.
Our method and its contributions can be summarized as follows:

\begin{itemize}

    \item 
    Text3DAug pioneers the use of generative models, prompting instances for augmentation.
    We evaluate the effectiveness of our novel pipeline through comprehensive experiments on LiDAR segmentation and detection benchmarks.

    \item
    Text3DAug does not require labels or trajectory information. Our instance engine is fully automated and generates a plethora of annotated instances without manual effort.
    This approach enables the scalability of our method, able to augment with potentially thousands of instances.

    \item
    Our method implements realistic placement and rendering of instances according to sensor characteristics. As such, Text3DAug is sensor agnostic, which we evaluate using various datasets.

    \item
    As a prompt-based method, Text3DAug is not constrained by dataset classes, and lends itself to the label-free training of new classes. We evaluate this with experiments on novel class discovery.

\end{itemize}

\section{RELATED WORK}
The natural imbalance in LiDAR point clouds requires extensive and varied datasets for deep learning. Simulated data has emerged as a viable alternative to real-world data acquisition. Additionally, data augmentation, including the particularly effective instance augmentation, has become a standard method for enhancing data diversity.

\subsection{Data Simulation}
Data simulation has emerged as a natural alternative to the time-consuming and expensive process of data acquisition and labeling.
Urban simulators such as SYNTHIA \cite{Synthia} and Carla \cite{Carla} are based on game engines, while others extend existing video games \cite{SimGTA-V, SimGTA-V-2}.
Simulators allow for data generation under various lighting and weather conditions, with differing dynamic object behaviour and new viewpoints, enabling the collection of diverse data for different sensor modalities.
Nonetheless, simulated scenes are created with significant manual effort for 3D asset creation, realistic placement, dynamic animation and rendering.
VirtualKITTI \cite{VirtualKITTI} and LiDARsim \cite{LiDARsim} instead use real world LiDAR scans to initialize digital twins, but again rely on manually labeled data in order to transfer object classes and positions. 

Despite its apparent benefits, a large domain gap remains between synthetic and real world data, resulting in a significant performance gap \cite{SynLiDAR, Image360, CosMIX}.
Currently, methods using synthetic data lag behind those using a modest amount of real labeled data and even further behind those using large scale datasets \cite{Image360}.
So called "sim2real" methods attempt to map real LiDAR characteristics to synthetic data \cite{SynLiDAR, LiDARsim, GANLidar} or mix real data into the training process \cite{CosMIX, Gipso}.
LiDAR-Aug \cite{fang2021lidar} inserts synthetic CAD models into real LiDAR pointclouds, followed by ray casting.
However, LiDAR-Aug requiring the costly curation or manual creation to obtain such models.
Moreover, CAD models themselves might vary in quality (poly count and detail) and in factors such as coordinate system definition.
For example, the mesh axis of a CAD model does not necessary align with the real center point, varying by object class and standard (e.g. ISO8855 \cite{ISO8855:2011} defining the rear axle of a car as the vehicle center), requiring post processing after curation.
Lastly, most simulation methods, including LiDAR-Aug do not account for the LiDAR-remission values absent from CAD models.

\subsection{Instance Augmentation}

Instance augmentation has been a significant step towards increasing 3D point cloud data diversity, especially of underrepresented classes, an aspect crucial for safety critical applications such as autonomous driving.
The pioneering work of Yan et al. \cite{yan2018second} laid the foundation, creating a database of "cut and paste" ground-truth instances for integration into LiDAR scans.
Zhou et al. \cite{zhou2021panoptic} extend this concept by oversampling rare class instances and adding local instance transformation in order to maximize data variance.
However, "cut and paste" methods preserve the point distribution and structure specific to the instances original position relative to the LiDAR sensor. Due to the radiating scanlines, the point density of an object decreases with increasing distance to the sensor.
Placing an instance closer or further from the sensor, results in a different point density when compared to its surroundings.
These issues, in combination with the random placement of instances, leads to unrealistic representations in a LiDAR scan.
Subsequent works \cite{hasecke2022can, RangeFormer} add some realism by removing points occluded by the added instances, based on the range-view representation of point clouds. However, depending on the grids angular resolution, the range-view representation can lead to data loss.

Real3DAug \cite{vsebek2022real3d} instead precomputes placement and occlusion maps to identify suitable scene positions, but with significant limitations. These maps are prohibitively time-consuming and computationally expensive. Because of this, the dataset is modified once prior to training, meaning that augmentation remains identical between epochs.
Besides trajectories and labelled instances, Real3DAug also requires further semantic labels for its maps, regardless of the LiDAR perception task, as its placement strategy differentiates between different ground types such as road and sidewalks.
During placement, instance orientation is computed based on estimated bounding boxes, an approximation since LiDAR data only covers the sensor-facing sides of objects \cite{LiDARsim}.



\begin{figure*}[htpb!]
  \setlength{\belowcaptionskip}{-12.5pt}
  \centering
  \includegraphics[width=2\columnwidth]{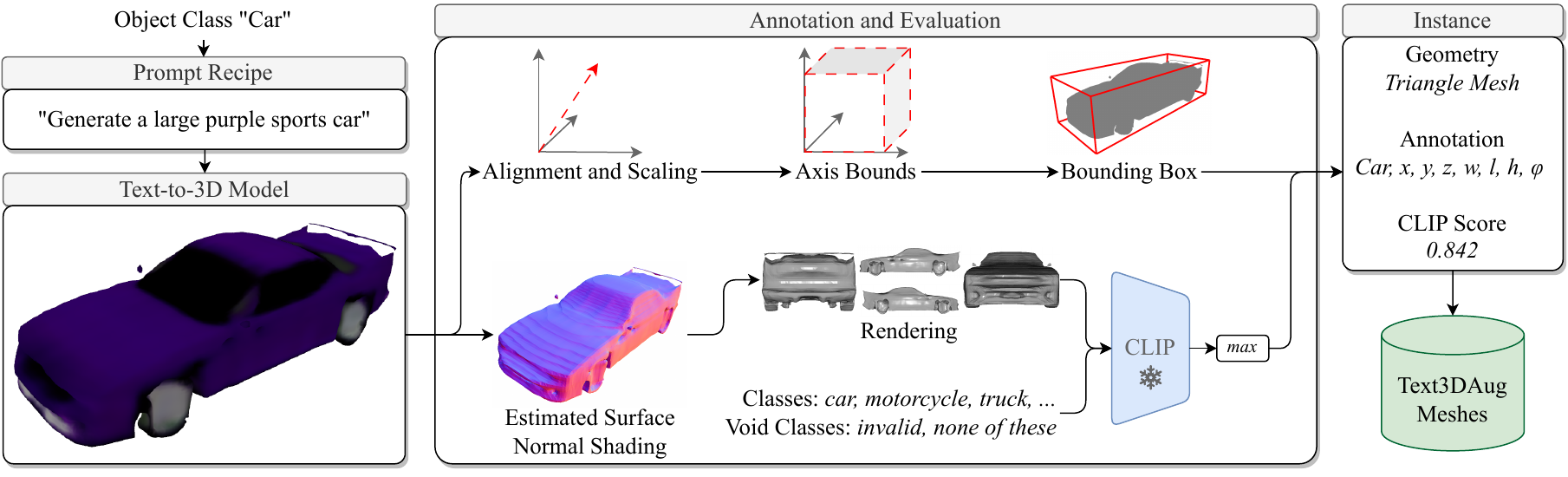}
  \caption{
  Our instance generation engine prompts text-to-3D models to generate mesh models. Annotations are derived from the mesh and CLIP scoring is used as a measure of quality. These are added to a database which will be used for the augmentation of LiDAR scans. The shown mesh model in this figure was generated by Shap-E \cite{Shape-E}.
  }
  \label{fig:Instance_Generation}
\end{figure*}

\section{APPROACH}

We identify three key observations regarding the current state-of-the-art in LiDAR instance augmentation. Firstly, "cut and paste" instance augmentation methods \cite{yan2018second, zhou2021panoptic, hasecke2022can, RangeFormer, vsebek2022real3d} depend on the availability of point-wise semantic and instance labels.
Secondly, the amount of instances available to these methods is limited by the size and variance of the dataset.
Thirdly, obtaining CAD models instead of instance cut-outs can be time consuming and expensive, with models potentially requiring post processing and varying in quality.

Thus, we propose Text3DAug, a fully automated and label free instance augmentation pipeline, the first leveraging generative models for 3D content creation.
Section \ref{Prompting} describes our standardized prompt recipe which we use to generate object meshes from text-to-3D models. These are then automatically post-processed, evaluated and labeled.
Our instance generation engine is described in Section \ref{Instance Generation} and in Figure \ref{fig:Instance_Generation}.
For augmentation, meshes are randomly selected, then placed and rendered as instances in LiDAR point clouds using the algorithms described in Section \ref{Placement}.
This process is further elaborated in Figure \ref{fig:Instance_Placement}.
Text3DAug is designed to be modular, allowing its components to be modified and extended with future research developments.

\subsection{Prompting}\label{Prompting}
Initially we explored the generation of meshes from text-to-3D models \cite{Point-E, Shape-E, Cap3D, GPT4Point, One2345} using prompts derived from the Q\&A output of a Large Language Model (LLM) \cite{GPT3.5}.
However, we observed that LLM generated prompts with multiple attributes such as "A man wearing a hat, walking and holding an umbrella" led to the generation of indistinguishable meshes across a variety of generative models.
In contrast, simple prompts such as "a man walking" produced plausible results.
Based on this observation and with efficiency in mind, we deemed the use of LLMs excessive and instead designed a fixed prompt recipe.
For a desired object class, we define a data store of synonyms and (if applicable) brand-names, e.g.  for the class car: \textit{sportscar}, \textit{convertible}, \textit{sedan}, \textit{SUV}, \textit{Ford}, etc.
From this, we build the prompt instruction and incorperate attributes like size and color to provide context.
Utilizing this recipe, we randomly sample prompts such as "Generate a large purple sports car" for the class car.

Our observation that simple prompts perform better is backed by current research, in that generative models struggle with prompts containing concepts or multiple attributes.
Specifically, this issue stems from coarse textual annotations in public datasets, with only a few attributes such as color and size \cite{Shape-E, Cap3D, GPT4Point}, or from drawbacks in the mesh optimization process resulting in imprecise geometries \cite{DreamFusion, RealFusion, One2345}.
With increased research and data in the space of generative models for 3D content, more complex prompt strategies will become possible.

\subsection{Instance Generation}\label{Instance Generation}
By employing the aforementioned prompt recipe, we utilize pre-trained text-to-3D models to generate object meshes for the desired classes.
These meshes are added to a database and later sampled for instance augmentation of the LiDAR scans. By pregenerating the meshes, we enable efficient augmentation during training.
A notable advantage of text-to-3D models is that meshes are generated with a specific orientation.
Contrary to methods such as Real3DAug \cite{vsebek2022real3d}, the orientation does not have to be estimated from partial LiDAR data.
We exploit this characteristic to derive precise bounding box annotations. This is achieved by axis aligning and transferring meshes into a common coordinate system, followed by fitting the bounding box based on the axis bounds.
Mesh vertices are scaled to a maximum height of one.
This later allows for the randomization of instance height during placement, during which the bounding box is transformed accordingly (refer to Section \ref{Placement}).
For the semantic label we assign the object class that was used for prompting.

The mesh quality is automatically evaluated based on the CLIP score \cite{CLIP}.
Since color information is not necessary for LiDAR data, we substitute mesh textures with shading derived from estimated surface normals, and then render the mesh from four views (front, back, both sides).
The removal of textures encourages CLIP to focus on shape.
CLIP similarity is determined for these four views in relation to the desired classes, as well as for the void classes such as \textit{none of these} or \textit{invalid}.
The highest score corresponding to the prompted class is assigned as the quality value. This approach is based on the understanding that certain classes have characteristic views that are more representative or identifiable. For example, a person is best recognized from the front, while a car might be best viewed from its side.
We evaluate the effectiveness of CLIP for this task in Section \ref{Table:MeshEval}.
Our instance generation engine is depicted in Figure \ref{fig:Instance_Generation}.

\subsection{Placement and Local Augmentation}\label{Placement}
Our Text3DAug pipeline introduces a systematic approach to instance placement in 3D point clouds. For further detail on this procedure we refer to Alg. \ref{Alg:Placement}. 
First, $n$ random object meshes are sampled from the previously created database $D$, according to the desired classes $C$, ensuring a diverse representation of object classes.
To add additional realism, we process the LiDAR data by mapping remission values according to their range, denoted as $R$. This approach enables the unsupervised acquisition of realistic remission values.
Remission values are assigned to the mesh vertices by sampling random values from $R$ corresponding to the range of each vertex relative to its position during placement.

Each mesh undergoes a random local transformation, which includes height scaling within a class-dependent appropriate range, followed by rotation, in order to reflect the stochastic nature of real-world data.
This is followed by a free-space analysis, which transforms a point cloud $P$ and the mesh's vertices into polar coordinates.
A random placement distance $r$ of the object to the sensor is chosen, and the relative azimuth span $\Delta \Phi(r)$ is derived from the mesh. 
Our algorithm filters the polar coordinates of the point cloud $P$ based on $\Delta \Phi$ and the mesh height, delineating viable regions.
Then, one of these regions is randomly chosen and the mesh is positioned within it.
If no regions are found, the process is repeated with a different object distance.
After a suitable region is found, we consider all points of $P$ above and below the mesh and take their minimum $z$-coordinate $z_{min}$ as the estimated ground level.
If no points are found, we also search within the area around the mesh, until $z_{min}$ is determined.
The mesh $z$-positioning is then corrected with $z_{min}$, anchoring it to the ground.
We found this approach to be particularly effective, enabling a seamless integration into urban landscapes, e.g. accurately placing objects on sidewalks or inclined roads.
Through this placement algorithm, meshes are inserted into the point cloud respecting its existing structure and spatial constraints.
For the last step of realistic placement, we calculate $\Phi_{min}$ and $\Phi_{max}$ for each vertical scanline (ring) and use these as a limit to remove points in front of the mesh or occluded by it.

\begin{algorithm}[ht]
\caption{Instance Augmentation of Text3DAug}\label{alg:instance_aug}
\begin{algorithmic}[1]

\State \textbf{Initialization:}
\State \hspace{\algorithmicindent} $D$: Database matching mesh paths to classes
\State \hspace{\algorithmicindent} $C$: List of dataset classes for augmentation
\State \hspace{\algorithmicindent} $R$: Remission file
\State \hspace{\algorithmicindent} Augmentation parameters ($n$, $w_{noise}$, $p_{drop}$)
\State \hspace{\algorithmicindent} LiDAR sensor parameters
\State \rule{\linewidth}{0.2pt} 

\State \textbf{Input:}
\State \hspace{\algorithmicindent} $P$: Point cloud
\State \hspace{\algorithmicindent} $L$: Annotations \Comment{\textcolor{gray}{bounding box or semantic label}}

\State
\State $C_n = $ Randomly select $n$ classes from $C$

\For{class $c$ in $C_n$}
    \State Select a random mesh from the class $c$ in $D$
    \State Add remission to mesh from $R$
    \State Random mesh height scaling and rotation
    \State Find viable regions in $P$ using free-space analysis
    \State Place mesh and adjust to estimated ground level $z_{min}$
    \State Render instance from mesh with ray casting
    \State Remove points behind or in front of instance
    \State Point removal with $p_{drop}$ and noise $w_{noise}$
    \State Add instance to $P$

    \If{Detection}
    \State Update $L$ according to instance placement
    \ElsIf{Segmentation}
    \State Update $L$ with class $c$ for instance
    \EndIf

\EndFor

\State \textbf{Return:}
\State \hspace{\algorithmicindent} Augmented point cloud $P_{aug}$ with instances
\State \hspace{\algorithmicindent} Updated annotations $L_{aug}$

\end{algorithmic}
\label{Alg:Placement}
\end{algorithm}



After placement, instances are rendered from the meshes using ray casting based on the LiDAR sensor parameters, crucial for simulating an authentic point distribution. Realism is enhanced by introducing noise with the weight $w_{noise}$ and point dropout with probability $p_{drop}$, emulating imperfections found in real-world data.
Annotations $L$ are adjusted accordingly, with the mesh class being assigned as the semantic label. For detection tasks, the bounding boxes of the instances are adjusted based on the instance's position, rotation, and height.
The augmented point cloud $P_{aug}$ and the updated annotations $L_{aug}$ are passed on for network training.

\section{EXPERIMENTS}

\begin{table*}[b]
\vspace*{-\baselineskip}
\centering
\caption{Evaluation comparing Text3DAug to other instance augmentation methods on the NuScenes \cite{kitti} detection task using PointPillars \cite{PointPillars}.
Classes \textit{trailer}, \textit{construction vehicle}, \textit{traffic cone} and \textit{barrier} were not augmented.
}
\resizebox{1.6\columnwidth}{!}{

\begin{tabular}{l@{}lccccccccccc}
\hline
\rowcolor[HTML]{EFEFEF} 
\multicolumn{2}{l}{\cellcolor[HTML]{EFEFEF}} 
& \rot{mAP$^{\text{\%}}$}        
& \rot{AP$_{\text{car}}^{\text{\%}}$   }   
& \rot{AP$_{\text{truck}}^{\text{\%}}$   } 
& \rot{AP$_{\text{bus}}^{\text{\%}}$    } 
& \rot{AP$_{\text{trailer}}^{\text{\%}}$ }
& \rot{AP$_{\text{construct.}}^{\text{\%}}$}
& \rot{AP$_{\text{pedestrian}}^{\text{\%}}$}
& \rot{AP$_{\text{motorcycle}}^{\text{\%}}$}
& \rot{AP$_{\text{bicycle}}^{\text{\%}}$  }
& \rot{AP$_{\text{traffic cone}}^{\text{\%}}$}
& \rot{AP$_{\text{barrier}}^{\text{\%}}$}
\\ \hline
                                & None (Baseline)  & 49.0          & 81.8          & 49.4          & 58.8          & 34.1          & 16.4                       & 74.6            & 47.5            & \textbf{23.0} & 50.2               & 53.9          \\
                                & Lidar-Aug* \cite{fang2021lidar}       & 49.2          & 82.3          & 47.1          & 59.2          & \textbf{35.8} & 16.5                       & \textbf{74.9}   & 51.0            & 19.7          & 51.0               & 54.2          \\
                                & GT-Aug \cite{yan2018second}           & 44.4          & 80.9          & 48.8          & \textbf{62.4} & 35.2          & 12.5                       & 72.0            & 31.2            & 6.2           & 46.0               & 48.4          \\
                                & Ours             & \textbf{49.4} & \textbf{82.4} & \textbf{49.5} & 58.7          & 34.3          & \textbf{16.8}              & 74.3            & \textbf{51.1}   & 20.4          & \textbf{51.6}      & \textbf{54.8} \\
\multirow{-5}{*}{
}   & Ours+GT-Aug    & 44.0          & 81.4          & 49.0          & 61.0          & 34.4          & 12.2                       & 72.0            & 30.7            & 5.2           & 45.3               & 48.6          \\ \hline
\multicolumn{13}{l}{*Re-implemented}                                                                                                                                                                                                                            
\end{tabular}

}
\label{Table:Detection_NuScenes}
\end{table*}

We conduct a series of experiments to assess the capabilities of our prompted Text3DAug pipeline and its components.
In Section \ref{Eval:Segmentation} and Section \ref{Eval:Detection}, we evaluate Text3DAug as an instance augmentation method for LiDAR perception, comparing it to relevant methods on the tasks of LiDAR segmentation and detection.
The label-free nature of our method lends itself to novel class discovery and we investigate whether text-generated meshes are sufficient for this task in Section \ref{Eval:NCD}.
Our choice of a generative text-to-3D model and CLIP scoring is examined in Section \ref{Mesh Comparison}.
Also, we investigate the scalability of our method through a trade-off between mesh quality and quantity in Section \ref{Quality vs Quantity}.
Finally, placement and local augmentation for realistic LiDAR rendering are assessed in Section \ref{Eval:Placement}.

For our experiments, we identify eight instance classes (\textit{car}, \textit{person}, \textit{bicycle}, \textit{bicyclist}, \textit{motorcycle}, \textit{motorcyclist}, \textit{truck}, and \textit{bus}) which are shared or can be remapped between the
SemanticKITTI \cite{semantic-kitti} and NuScenes semantic segmentation datasets \cite{nuscenes_seg},
as well as the KITTI \cite{kitti} and NuScenes \cite{nuscenes} detection datasets. We also use these for our evaluation.
Unless otherwise stated, we use the original implementations for generative models, segmentation models and augmentation pipelines. Detection experiments are based on the OpenPCDet \cite{OpenPCDet} framework.
We use CLIP \cite{CLIP} with ViT-L/14 \cite{ViT} for mesh evaluation.
Training configurations and settings will be made available along with our code release.

\subsection{Segmentation}\label{Eval:Segmentation}

\begin{table}[t]
\centering
\caption{Comparison of instance augmentation methods for LiDAR semantic segmentation mIoU.
}

\resizebox{1\columnwidth}{!}{

\begin{tabular}{lccccc}
\hline
\rowcolor[HTML]{EFEFEF} 
                                                        &                                                                               &
                                                    \multicolumn{2}{c}{\cellcolor[HTML]{EFEFEF}\begin{tabular}[c]{@{}c@{}}SemanticKITTI \cite{semantic-kitti}\end{tabular}}  & \multicolumn{2}{c}{\cellcolor[HTML]{EFEFEF}\begin{tabular}[c]{@{}c@{}}NuScenes \cite{nuscenes_seg}\end{tabular}}    \\
\rowcolor[HTML]{EFEFEF} 
Method                                                  & Type                                                                          & \begin{tabular}[c]{@{}c@{}}C3D \\ \cite{Cylinder3D}\end{tabular}  & \begin{tabular}[c]{@{}c@{}}SPVCNN \\ \cite{SPVCNN}\end{tabular} &\begin{tabular}[c]{@{}c@{}}C3D \\ \cite{Cylinder3D}\end{tabular} & \begin{tabular}[c]{@{}c@{}}SPVCNN \\ \cite{SPVCNN}\end{tabular} \\ \hline
None (Baseline)                                         & \multicolumn{1}{c|}{-}                                                        & 61.70                                             & \multicolumn{1}{c|}{63.72}                           &75.74                                            & 69.50                                               \\
Lidar-Aug* \cite{fang2021lidar}                                              & \multicolumn{1}{c|}{CAD}                                                      & 63.47                                             & \multicolumn{1}{c|}{64.40}                           & 75.49                                             &69.58\\
Real3DAug \cite{vsebek2022real3d}                                              & \multicolumn{1}{c|}{Labeled}                                                  & 63.25                                             & \multicolumn{1}{c|}{65.00}                            & -                                                & -                                                   \\
GT-Aug \cite{yan2018second}                                                  & \multicolumn{1}{c|}{Labeled}                                                  & {66.07}                                       & \multicolumn{1}{c|}{65.33}                           & 75.05                                            & 67.96                                               \\
Ours                                                    & \multicolumn{1}{c|}{Text}                                                     & 64.80                                             & \multicolumn{1}{c|}{65.18}                           & \textbf{75.98}                                   & {\textbf{69.85}}                                        \\
\begin{tabular}[c]{@{}l@{}}Ours\\+GT-Aug\end{tabular} & \multicolumn{1}{c|}{\begin{tabular}[c]{@{}c@{}}Text\\ + Labeled\end{tabular}} & \textbf{66.94}                                    & \multicolumn{1}{c|}{\textbf{67.52}}                  & 74.80                                            & 68.78                                               \\ \hline
\multicolumn{6}{l}{* Re-implemented}                                                                                                                                                                                                                                                                                   
\end{tabular}

}
\vspace{-\baselineskip}
\vspace{-\baselineskip}
\label{Table:Segmentation}
\end{table}
We evaluate our method Text3DAug on the SemanticKITTI and NuScenes semantic segmentation datasets using the state-of-the-art networks SPVCNN \cite{SPVCNN} and Cylinder3D \cite{Cylinder3D}. For all methods we place five instances from 1,000 available meshes per class into each scan, generated by Shap-E \cite{Shape-E}.
Results are shown in Table \ref{Table:Segmentation}.

In our study on the SemanticKITTI dataset, all instance augmentation methods led to improvements. GT-Aug, a representative "cut and paste" approach with random placement, performs well, but is constrained by a limited pool of instances from the dataset with under-represented classes.
Since Text3DAug operates independently of dataset content, its scalability can address this issue.
Combining both GT-Aug and Text3DAug yielded the best results, indicating a synergistic effect and further improving model performance. Using SPVCNN, Text3DAug performed almost equal to GT-Aug.
The worse results in Table \ref{Table:Segmentation} of Real3DAug can be attributed to the fact, that instance placement is precomputed prior to training, not between epochs.
LiDAR-Aug does not consider the remission values of the CAD models, which has a negative effect on network performance (see Section \ref{Eval:Placement}).

Unexpectedly, GT-Aug was detrimental to network performance on NuScenes.
One downside of the "cut and paste" approach is that instances retain properties of their original location.
The sensor used for the data acquisition of NuScenes has less scanlines and a different field of view, leading to a significantly reduced point density when compared to the sensor used for KITTI.
The combination of this specific sensor and the random instance placement may result in an unrealistic augmentation that negatively impacts network performance. 
As such, GT-Aug does not generalize well between these two datasets.
Generally, improvements were less pronounced compared to those on the SemanticKITTI dataset, which can be attributed to the diversity of the NuScenes dataset.
NuScenes has significantly more LiDAR scans and a greater variety in their 1,000 short scenes compared to SemanticKITTIs 22 long scenes. Furthermore NuScenes has ca. $4.4\times$ as many instances \cite{nuscenes_seg}.
Considering the substantial differences between the datasets, it is logical that instance augmentation has less of an impact.

In summary, Text3DAug leads to improvements over the baseline in all cases, synergizing with GT-Aug for KITTI and leading to the best results. For NuScenes, Text3DAug performs the best among all methods. In general, our standalone method performs on par with or better than established state-of-the-art methods, albeit completely label free.

\subsection{Detection}\label{Eval:Detection}

\begin{table}[t]
\centering
\caption{Evaluation comparing our method Text3DAug to other instance augmentation methods on the KITTI \cite{kitti} detection task.
}
\resizebox{1\columnwidth}{!}{

\begin{tabular}{@{}l@{}l@{\hspace{0.1cm}}c@{\hspace{0.2cm}}c@{\hspace{0.2cm}}c@{\hspace{0.3cm}}c@{\hspace{0.2cm}}c@{\hspace{0.2cm}}c@{\hspace{0.3cm}}c@{\hspace{0.2cm}}c@{\hspace{0.2cm}}c}
\rowcolor[HTML]{EFEFEF} \hline
\multicolumn{2}{l}{\cellcolor[HTML]{EFEFEF}}                         & \multicolumn{3}{c}{\cellcolor[HTML]{EFEFEF}AP$_{\text{car}}^{\text{70\%}}$}                                                                   & \multicolumn{3}{c}{\cellcolor[HTML]{EFEFEF}AP$_{\text{pedestrian}}^{\text{50\%}}$}                                                            & \multicolumn{3}{c}{\cellcolor[HTML]{EFEFEF}AP$_{\text{cyclist}}^{\text{50\%}}$}                                                                \\
\rowcolor[HTML]{EFEFEF} 
\multicolumn{2}{l}{\multirow{-2}{*}{\cellcolor[HTML]{EFEFEF}Method}} & easy                                  & mod.                                  & hard                                  & easy                                  & mod.                                  & hard                                  & easy                                  & mod.                                  & hard                                  \\ \hline
                                         &None (Baseline)           &{89.11}          &{79.24}          &{78.58}          &{{66.72}}    &{{60.23}}    &{{58.47}}    &{76.93}          &{57.58}          &{56.55}          \\
                                         &Lidar-Aug* \cite{fang2021lidar}                &{88.89}          &{79.08}          &{78.57}          &{61.36}          &{58.96}          &{56.77}          &{83.92}          &{64.78}          &{58.49}          \\
                                         &GT-Aug \cite{yan2018second}                    &{{89.25}}    &{{83.27}}    &{{78.78}}    &{64.79}          &{58.32}          &{54.20}           &{{86.13}}    &{{71.96}}    &{{68.50}}     \\
                                         &Ours                      &{88.80}           &{78.94}          &{78.29}          &{\textbf{67.17}} &{\textbf{60.57}} &{\textbf{58.92}} &{85.60}           &{65.05}          &{63.71}          \\
\multirow{-5}{*}{\rot{\begin{tabular}{c}
  PV-RCNN \\
  \cite{PV-RCNN}
\end{tabular}}}                &Ours+GT-Aug             & \textbf{89.75} & \textbf{83.78} & \textbf{78.98} & 65.76          & 59.53          & 54.79          & \textbf{93.37} & \textbf{73.84} & \textbf{70.16} \\ \hline
                                         &None (Baseline)           & 86.39                                 & 75.50                                  & 72.64                                 & 52.04                                 & 47.20                                  & 44.81                                 & 65.03                                 & 49.61                                 & 45.76                                 \\
                                         &Lidar-Aug* \cite{fang2021lidar}                & 86.54                                 & 76.10                                  & 73.06                                 & 53.10                                  & 47.81                                 & 44.83                                 & 65.28                                 & 51.94                                 & 49.13                                 \\
                                         &GT-Aug \cite{yan2018second}                    & {88.06}                           & \textbf{78.50}                         & {77.10}                            & \textbf{56.40}                         & \textbf{53.12}                        & \textbf{48.33}                        & {82.42}                           & {64.04}                           & {60.88}                           \\
                                         &Ours                      & 86.51                                 & 75.78                                 & 72.75                                 & 53.74                                 & {50.82}                           & {47.32}                           & 70.62                                 & 57.61                                 & 53.20                                  \\
\multirow{-5}{*}{\rot{\begin{tabular}{c}
  SECOND \\
  \cite{yan2018second}
\end{tabular}}}                 &Ours+GT-Aug             & \textbf{88.40}                         & {78.40}                            & \textbf{77.13}                        & {54.71}                           & 50.12                                 & 45.47                                 & \textbf{82.44}                        & \textbf{66.26}                        & \textbf{62.03}                        \\ \hline
                                         &None (Baseline)           & 84.24                                 & 72.51                                 & 67.35                                 & 46.69                                 & 43.49                                 & 40.47                                 & 62.84                                 & 44.13                                 & 41.88                                 \\
                                         &Lidar-Aug* \cite{fang2021lidar}                & 86.33                                 & 73.54                                 & 67.92                                 & 40.06                                 & 36.77                                 & 35.24                                 & 62.52                                 & 42.88                                 & 40.96                                 \\
                                         &GT-Aug \cite{yan2018second}                   & {86.43}                           & {77.24}                           & {75.28}                           & \textbf{55.79}                        & \textbf{50.94}                        & \textbf{46.66}                        & \textbf{82.56}                        & \textbf{63.68}                        & \textbf{60.79}                        \\
                                         &Ours                      & 86.39                                 & 73.91                                 & 68.49                                 & 40.93                                 & 37.88                                 & 35.59                                 & 56.89                                 & 43.81                                 & 41.65                                 \\
\multirow{-5}{*}{\rot{\begin{tabular}{c}
  PointPillar \\
  \cite{PointPillars}
\end{tabular}}}            & Ours+GT-Aug             & \textbf{87.65}                        & \textbf{77.48}                        & \textbf{75.59}                        & {54.85}                           & {49.76}                           & {46.35}                           & {78.33}                           & {61.91}                           & {58.92}                           \\ \hline
\multicolumn{11}{l}{* Re-implemented}                                                                                                                                                                                                                                                                                                                                                                                                       
\end{tabular}

}
\vspace{-\baselineskip}
\vspace{-\baselineskip}
\label{Table:Detection_KITTI}
\end{table}


\begin{table*}[t]
\centering
\caption{
Evaluation of generative models in our instance generation engine.
We compare the segmentation mIoU for all $19$ training classes
and the mIoU for augmented instance classes only on SemanticKITTI \cite{semantic-kitti}.
}
\resizebox{2\columnwidth}{!}{

\begin{tabular}{lcccccccc}
\hline
\rowcolor[HTML]{EFEFEF} 
\multicolumn{2}{l}{\cellcolor[HTML]{EFEFEF}Generative Model}                      & \begin{tabular}[c]{@{}c@{}}None\\ (Baseline)\end{tabular} & Point-E \cite{Point-E} & Shap-E \cite{Shape-E}         & \begin{tabular}[c]{@{}c@{}}One-2-3-45 \cite{One2345}\\ (SDXL-T \cite{SDXL-Turbo})\end{tabular} & \begin{tabular}[c]{@{}c@{}}Cap3D \cite{Cap3D}\\ (Point-E)\end{tabular} & \begin{tabular}[c]{@{}c@{}}Cap3D \\ (Shap-E)\end{tabular} & Gpt4Point \cite{GPT4Point} \\ \hline
\multicolumn{2}{l}{\begin{tabular}[c]{@{}l@{}}All class mIoU\end{tabular}}      & 61.71                                                     & 62.29   & \textbf{63.93} & 62.58                                                         & 63.53                                                     & 63.88                                                    & 63.12     \\
\multicolumn{2}{l}{\begin{tabular}[c]{@{}l@{}}Instances mIoU\end{tabular}}  & 59.10                                                     & 61.46   & \textbf{64.49} & 60.66                                                         & 62.91                                                     & 64.15                                                    & 63.03     \\
\multicolumn{2}{l}{\begin{tabular}[c]{@{}l@{}}Instances CLIP\end{tabular}} & -                                                         & 0.585   & \textbf{0.588} & 0.454                                                         & 0.555                                                     & 0.575                                                    & 0.552    \\ \hline \hline
                                                      & IoU                        & 68.33                                                     & 69.82   & 67.89          & 67.12                                                         & 72.79                                                     & 73.09                                                    & 72.29     \\
\multirow{-2}{*}{Person}                              & CLIP                       & -                                                         & 0.355   & 0.357          & 0.426                                                         & 0.436                                                     & 0.257                                                    & 0.389     \\ \hline
                                                      & IoU                        & 86.13                                                     & 89.25   & 82.77          & 85.55                                                         & 88.00                                                     & 86.15                                                    & 84.96     \\
\multirow{-2}{*}{Bicyclist}                           & CLIP                       & -                                                         & 0.827   & 0.825          & 0.435                                                         & 0.750                                                     & 0.779                                                    & 0.467     \\ \hline
                                                      & IoU                        & 47.79                                                     & 40.97   & 45.40          & 45.47                                                         & 44.68                                                     & 45.41                                                    & 46.84     \\
\multirow{-2}{*}{Bicycle}                             & CLIP                       & -                                                         & 0.451   & 0.444          & 0.322                                                         & 0.467                                                     & 0.479                                                    & 0.777     \\ \hline
                                                      & IoU                        & 1.51                                                      & 12.68   & 37.85          & 0.03                                                          & 9.11                                                      & 37.17                                                    & 26.35     \\
\multirow{-2}{*}{Motorcyclist}                        & CLIP                       & -                                                         & 0.268   & 0.287          & 0.131                                                         & 0.395                                                     & 0.299                                                    & 0.415     \\ \hline
                                                      & IoU                        & 63.73                                                     & 45.25   & 60.16          & 51.79                                                         & 62.79                                                     & 60.73                                                    & 62.48     \\
\multirow{-2}{*}{Motorcycle}                          & CLIP                       & -                                                         & 0.593   & 0.611          & 0.268                                                         & 0.480                                                     & 0.560                                                    & 0.495     \\ \hline
                                                      & IoU                        & 96.06                                                     & 96.38   & 94.51          & 95.78                                                         & 96.57                                                     & 95.28                                                    & 95.38     \\
\multirow{-2}{*}{Car}                                 & CLIP                       & -                                                         & 0.813   & 0.787          & 0.805                                                         & 0.776                                                     & 0.871                                                    & 0.761     \\ \hline
                                                      & IoU                        & 44.26                                                     & 54.74   & 51.41          & 52.86                                                         & 52.16                                                     & 40.71                                                    & 49.23     \\
\multirow{-2}{*}{Bus}                                 & CLIP                       & -                                                         & 0.249   & 0.254          & 0.599                                                         & 0.115                                                     & 0.288                                                    & 0.110     \\ \hline
                                                      & IoU                        & 64.99                                                     & 85.61   & 75.93          & 86.66                                                         & 77.20                                                     & 74.66                                                    & 66.72     \\
\multirow{-2}{*}{Truck}                               & CLIP                       & -                                                         & 0.582   & 0.576          & 0.385                                                         & 0.380                                                     & 0.535                                                    & 0.372     \\  \hline
\end{tabular}

}
\label{Table:MeshEval}
\vspace*{-\baselineskip}
\end{table*}
\begin{table}[b!]
\vspace*{-\baselineskip}
\vspace*{-0.5\baselineskip}
\caption{
Novel class discovery using Text3DAug for KITTI \cite{kitti} object detection with SECOND \cite{yan2018second}.
}
\begin{subtable}{1\columnwidth}
\centering

\resizebox{1\columnwidth}{!}{

\begin{tabular}{lcccccc}
\hline
\rowcolor[HTML]{EFEFEF} 
\cellcolor[HTML]{EFEFEF}                              & \multicolumn{3}{c}{\cellcolor[HTML]{EFEFEF}AP$_{\text{pedestrian}}^{\text{50\%}}$} & \multicolumn{3}{c}{\cellcolor[HTML]{EFEFEF}AP$_{\text{cyclist}}^{\text{50\%}}$} \\
\rowcolor[HTML]{EFEFEF} 
\multirow{-2}{*}{\cellcolor[HTML]{EFEFEF}Method} & easy               & mod               & hard              & easy            & mod             & hard            \\ \hline
Fully-Supervised \cite{yan2018second}                                             & 52.04               & 47.20              & 44.81              & 65.03            & 49.61            & 45.76            \\
Novel Class Discovery (ours)                                                  & 28.14               & 27.07              & 24.02              & 12.16            & 10.50            & 10.59            \\ \hline
\end{tabular}

}
\end{subtable}

\label{Table:NCD-Detection}

\vspace{1em}
\caption{
Novel class discovery using Text3DAug for SemanticKITTI \cite{semantic-kitti} segmentation with Cylinder3D \cite{Cylinder3D}.
}
\begin{subtable}{1\columnwidth}

\centering

\resizebox{0.7\columnwidth}{!}{

\begin{tabular}{lcc}
\hline
\rowcolor[HTML]{EFEFEF} 
\cellcolor[HTML]{EFEFEF}                              & \multicolumn{2}{c}{\cellcolor[HTML]{EFEFEF}max. class IoU} \\
\rowcolor[HTML]{EFEFEF} 
\multirow{-2}{*}{\cellcolor[HTML]{EFEFEF}Method} & Car                      & Pedestrian                      \\ \hline
Fully-Supervised \cite{Cylinder3D}                                             & 96.09                            & 75.71                     \\
Novel Class Discovery (ours)                                                  & 49.00                            & 16.80                     \\ \hline
\end{tabular}

}
\end{subtable}

\label{Table:NCD-Segmentation}

\end{table}

We evaluate 3D object detection performance on the KITTI and NuScenes datasets. We use the detectors PV-RCNN \cite{PV-RCNN}, SECOND \cite{yan2018second} and PointPillar \cite{PointPillars}  for KITTI and PointPillar for NuScenes. Following the KITTI benchmark convention, the mean Average Precision (mAP) for pedestrians and cyclists is calculated with a $50\%$ overlap, while for cars it is calculated with a $70\%$ overlap. As for semantic segmentation, the same 1,000 meshes generated by Shap-E \cite{Shape-E} are used for each class, with five being randomly inserted per scan.

Table \ref{Table:Detection_NuScenes} shows that Text3DAug leads to an improvement over the baseline. Similar to the segmentation results in Table \ref{Eval:Segmentation}, Text3DAug best complements GT-Aug on KITTI detection.
However, NuScenes results show the same negative effect for GT-Aug, in which the combination of "cut and paste" instances with data of this specific sensor results in an unrealistic augmentation. Here GT-Aug is detrimental to NuScenes detection performance, a problem that does not affect our method.
In the case on NuScenes, Text3DAug performs best as a standalone method. The results on both datasets complement previous experiments, highlighting the effectiveness of Text3DAug across sensors and tasks.

\subsection{Novel Class Discovery}\label{Eval:NCD}
Due to its prompt-based nature, Text3DAug lends itself to novel class discovery. We perform experiments for both detection and segmentation, comparing fully-supervised network performance to performance learning a class without any real labeled data.
To achieve this, we  remove LiDAR points and labels of a specific class in the training split of a dataset and instead placing Text3DAug instances of this class during training.

The evaluation split remains unchanged, with the respective metric serving as the benchmark for class learning.
The aim of these experiments is to determine if a network can distinguish a new class in real data using only text prompts in Text3DAug. 
Cylinder3D \cite{Cylinder3D} is used for SemanticKITTI \cite{semantic-kitti} segmentation, while SECOND \cite{yan2018second} is used for KITTI \cite{kitti} detection.
The results are depicted in Tables \ref{Table:NCD-Segmentation} and \ref{Table:NCD-Detection}. Without any labels, Text3DAug achieves notable results, showing that new classes can be learned just from text. These results are demonstrate the significant potential of Text3DAug, especially when compared to the baseline using fully-labeled data. This opens future paths of research, integrating Text3DAug into label-efficient or unsupervised methods.
A downside of removing existing classes in this evaluation, is that empty regions are left in the LiDAR scan, potentially leading to worse results.

\subsection{Instance Generation}\label{Mesh Comparison}
We examine the influence of various state-of-the-art mesh generation models on the performance of Text3DAug.
To ensure a fair comparison, we use identical prompts and generate 250 meshes per class (without CLIP filtering). 
We choose LiDAR semantic segmentation as the measure of effectiveness, as fine grained point-level classification is required. Experiments are conducted on the SemanticKITTI dataset using Cylinder3D \cite{Cylinder3D} for segmentation. We add five instances using Text3DAug during training, which are randomly chosen from the eight instance classes (\textit{car}, \textit{person}, \textit{bicycle}, \textit{bicyclist}, \textit{motorcycle}, \textit{motorcyclist}, \textit{truck}, and \textit{bus}).

For Point-E \cite{Point-E}, Shap-E \cite{Shape-E}, Cap3D \cite{Cap3D} and GPT4Point \cite{GPT4Point} we generate the out-of-domain classes \textit{motorcyclist} and \textit{bicyclist} by combining the meshes of \textit{person} with those of \textit{motorcycle} or \textit{bicycle} together. This limitation to these classes does not affect One-2-3-45 \cite{One2345} and as such we include both classes in the evaluation. We determined a failure rate of $23.2\%$ for One-2-3-45 stemming from an unstable optimization process, in such cases requiring re-prompting with new prompts. Results are listed in Table \ref{Table:MeshEval}.

Independent of the generative model used, our method improves performance over the baseline, with Shap-E showing the best results.
Notably, combining \textit{person} with \textit{motorcycle} meshes to create the out-of-domain class \textit{motorcyclist} proved effective.
Furthermore, no single generative model excels across all classes; rather, considerable variation remains in which model performs best for a specific class. For the classes \textit{bicycle} and \textit{motorcycle}, no improvement was observed compared to the baseline.
This variability and existence of out-of-domain classes indicates the potential remaining in the field of 3D content generation.
As our Text3DAug is not reliant on one single generative model, we hypothesize that Text3DAug will synergize well with future research as such issues are resolved.

We also evaluate the average CLIP score of all meshes in comparison to the resulting performance.
There is a moderate positive correlation of $R = 0.69$ between the CLIP score and resulting mIoU.
Although the sample size of six models may be too small to establish a reliable correlation value, it does suggest a discernible trend: the highest CLIP score aligns with the best performing model measured by mIoU.

\subsection{Placement and Local Augmentation}\label{Eval:Placement}
The experiments in this section cover the design choices for the realistic placement and rendering of meshes in LiDAR scans as instances (described in Section \ref{Placement}). For this we use Cylinder3D \cite{Cylinder3D} for LiDAR segmentation on SemanticKITTI \cite{semantic-kitti} with 250 random meshes  per class for Text3DAug.

\noindent\begin{minipage}{\columnwidth} 
\vspace{0.5\baselineskip}
\captionof{table}{
Evaluation of the placement algorithm and remission sampling for our Text3DAug.}
\centering
\resizebox{0.5\columnwidth}{!}{

\begin{tabular}{lcc}
\hline
\rowcolor[HTML]{EFEFEF} 
Placement & Remission & mIoU           \\ \hline
Random    & \checkmark       & 63.62          \\
Realistic & \checkmark       & \textbf{63.93} \\
Realistic &         & 62.49          \\ \hline
\end{tabular}
}
\label{Table:Placement_and_Remission}
\vspace{0.5\baselineskip}

  \centering
  \includegraphics[width=1\columnwidth]{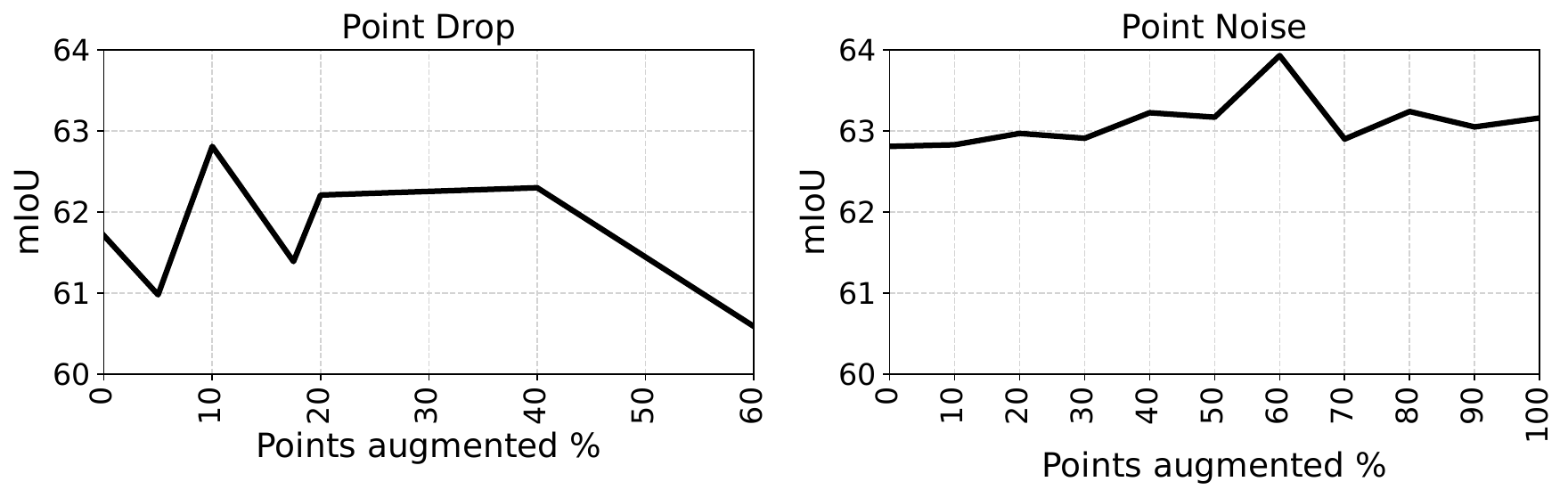}

  \captionof{figure}{
  Impact of removing points and adding noise on Text3DAug instances on SemanticKITTI \cite{semantic-kitti} semantic segmentation performance.
  }
  \label{fig:LocalAug}

\end{minipage}

First, we evaluate our choice of point drop and refer to Figure \ref{fig:LocalAug}. We find that removing points with a probability of $10\%$ from added instances performs best. Using this value, we then assess point noise, determining that affecting $60\%$ of points with noise further improves results.
Then, we evaluate our placement algorithm and the addition of remission values to meshes. We find that randomly placing instances performs worse and that remission values are critical for a network. Results are depicted in Table \ref{Table:Placement_and_Remission}, with the combination of realistic placement and remission values performing best.

\begin{figure}[]
  \centering
  \includegraphics[width=0.9\columnwidth]{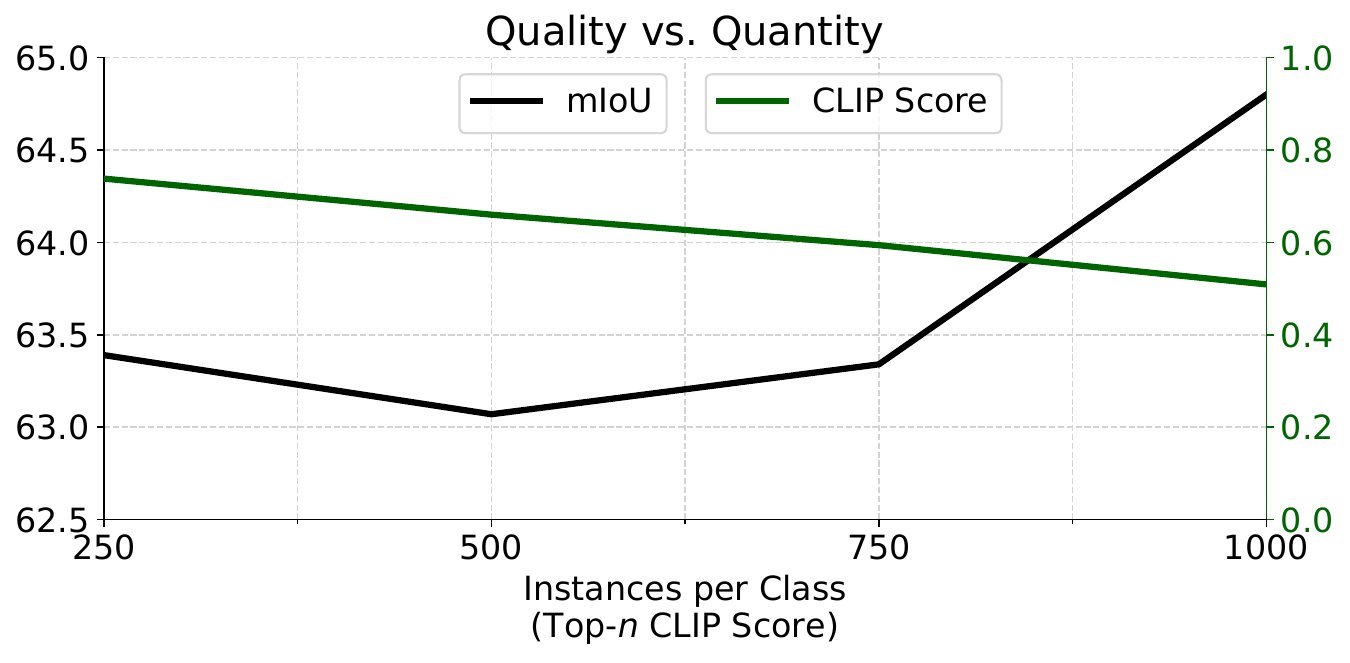}
  \caption{
We assess the impact of quality versus quantity on our pipeline, filtering the meshes by CLIP score and comparing results for the segmentation mIoU on SemanticKITTI \cite{semantic-kitti}.
  }
  \label{fig:Quality-vs-Quantity}
\vspace*{-\baselineskip}
\end{figure}

\subsection{Quality vs. Quantity}\label{Quality vs Quantity}
In this section, we investigate whether quality or quantity of meshes is more crucial for Text3DAug. For this, we extend the Shap-E database of meshes to 1,000 meshes per class. We then filter available meshes using the CLIP score as a measure of quality, selecting the top 250, 500 and 750 meshes for random sampling, in contrast to using all 1,000. Experiments are conducted using Cylinder3D \cite{Cylinder3D} on SemanticKITTI. We add five instances using Text3DAug during training. Results are plotted in Figure \ref{fig:Quality-vs-Quantity}, showing that the amount of available meshes for instance placement has a greater impact than the hypothetical quality based on the CLIP score.
This showcases the benefit of scalability provided by our method.
Notably, with 250 meshes filtered by CLIP score, Text3DAug performs slightly worse compared to the 250 random meshes from Section \ref{Mesh Comparison}.
These results contrast the moderate correlation of CLIP / mIoU from Section \ref{Mesh Comparison}, leading us to the conclusion that a diversity of meshes is most important for resulting model performance. Low quality meshes possibly have a regularizing effect, in turn leading to better data variance.

\section{CONCLUSION}
In this work, we propose Text3DAug, exploiting the recent advent of generative text-to-3D models for LiDAR instance augmentation.
Current instance augmentation methods rely on large datasets, labels or manual effort. Tackling these limitation, we design an automatic engine to generate meshes and annotations from text prompts.
These are then realistically placed into LiDAR scans and rendered as instances according to the sensor characteristics.
The number of instances available to our method is not limited by dataset size, offering a practical and scalable approach.
We show the effectiveness of Text3DAug through comprehensive evaluation on the tasks of LiDAR semantic segmentation and detection, outperforming or performing on-par with comparable methods, however without their drawbacks.
We further experiment with Text3DAug for novel class discovery, demonstrating the capability to learn new classes solely from text, potentially enhancing label-efficient or unsupervised research.
In future work, our modular pipeline can accommodate new generative networks or prompting techniques, developing with progress in the state of the art.
Furthermore, we see potential in expanding Text3DAug to different sensor modalities such as radar \cite{RadarPillars} and various methods of depth sensing.


\section*{ACKNOWLEDGMENT}
This work was funded by the German Federal Ministry for Economic Affairs and Climate Action (BMWK) under the grant AuReSi (KK5335501JR1) and SERiS (KK5335502LB3).

\bibliographystyle{./IEEEtran}
\bibliography{./IEEEabrv,./sources}

\end{document}